%% file: Article.tex
\documentclass[times, twoside]{zHenriquesLab-StyleBioRxiv}

\usepackage{dsfont}
\usepackage{blindtext}
\usepackage{xcolor, subfig, booktabs}

\newcommand{\bunderbrace}[2]{%
\begin{array}[t]{@{}c@{}}
\underbrace{#1}\\
#2
\end{array}
}

\leadauthor{Bala} 

\begin{document}
\title{Self-supervised Secondary Landmark Detection via 3D Representation Learning}
\shorttitle{}
\author[1]{Praneet C. Bala}
\author[2,*]{Jan Zimmermann}
\author[1,*,\Letter]{Hyun Soo Park}
\author[2,*]{Benjamin Y. Hayden}

\affil[1]{Department of Computer Science and Engineering, University of Minnesota, Minneapolis MN 55455 }
\affil[2]{Department of Neuroscience, Center for Magnetic Resonance Research, University of Minnesota, Minneapolis MN 55455}
\affil[*]{indicates co-last authors}
\maketitle

\vspace{-5mm}
\begin{abstract}
Recent technological developments have spurred great advances in the computerized tracking of joints and other landmarks in moving animals, including humans. Such tracking promises important advances in biology and biomedicine. Modern tracking models depend critically on labor-intensive annotated datasets of primary landmarks by non-expert humans. However, such annotation approaches can be costly and impractical for secondary landmarks, that is, ones that reflect fine-grained geometry of animals, and that are often specific to customized behavioral tasks. Due to visual and geometric ambiguity, non-experts are often not qualified for secondary landmark annotation, which can require anatomical and zoological knowledge. These barriers significantly impede downstream behavioral studies because the learned tracking models exhibit limited generalizability. We hypothesize that there exists a shared representation between the primary and secondary landmarks because the range of motion of the secondary landmarks can be approximately spanned by that of the primary landmarks. We present a method to learn this spatial relationship of the primary and secondary landmarks in three dimensional space, which can, in turn, self-supervise the secondary landmark detector. This 3D representation learning is generic, and can therefore be applied to various multiview settings across diverse organisms, including macaques, flies, and humans. 
\end {abstract}

\begin{corrauthor}
Hyun Soo Park, hspark\at umn.edu
\end{corrauthor}


\vspace{-5mm}
\input{01_Introduction}
\input{02_Result}
\input{04_Discussion}
\input{03_Method}

\input{05_Bibliography}
\end{document}

%% file: 01_Introduction.tex
\section*{Introduction}
Automated identification and tracking of important joints or other body landmarks has become an important method in biology and biomedicine. These tracking approaches have been leveraged by modern computer vision models that are designed to learn the complex visual and geometric relationships of landmarks from large annotated datasets~\cite{wei2016cpm,newell:2016,toshev:2014,cao2017realtime,fang2017rmpe, Mathis2018, Bala2020, gunel:2019,li:2020}. As a result, it is currently possible to computationally analyze the behaviors of many animals, including humans~\cite{wei2016cpm,newell:2016,toshev:2014,cao2017realtime,fang2017rmpe}, mice~\cite{Mathis2018}, monkeys~\cite{Bala2020}, and flies~\cite{gunel:2019,li:2020} without the use of specialized markers in a variety of contexts.


\begin{figure*}[t]
\begin{center}
\includegraphics[width=1\textwidth]{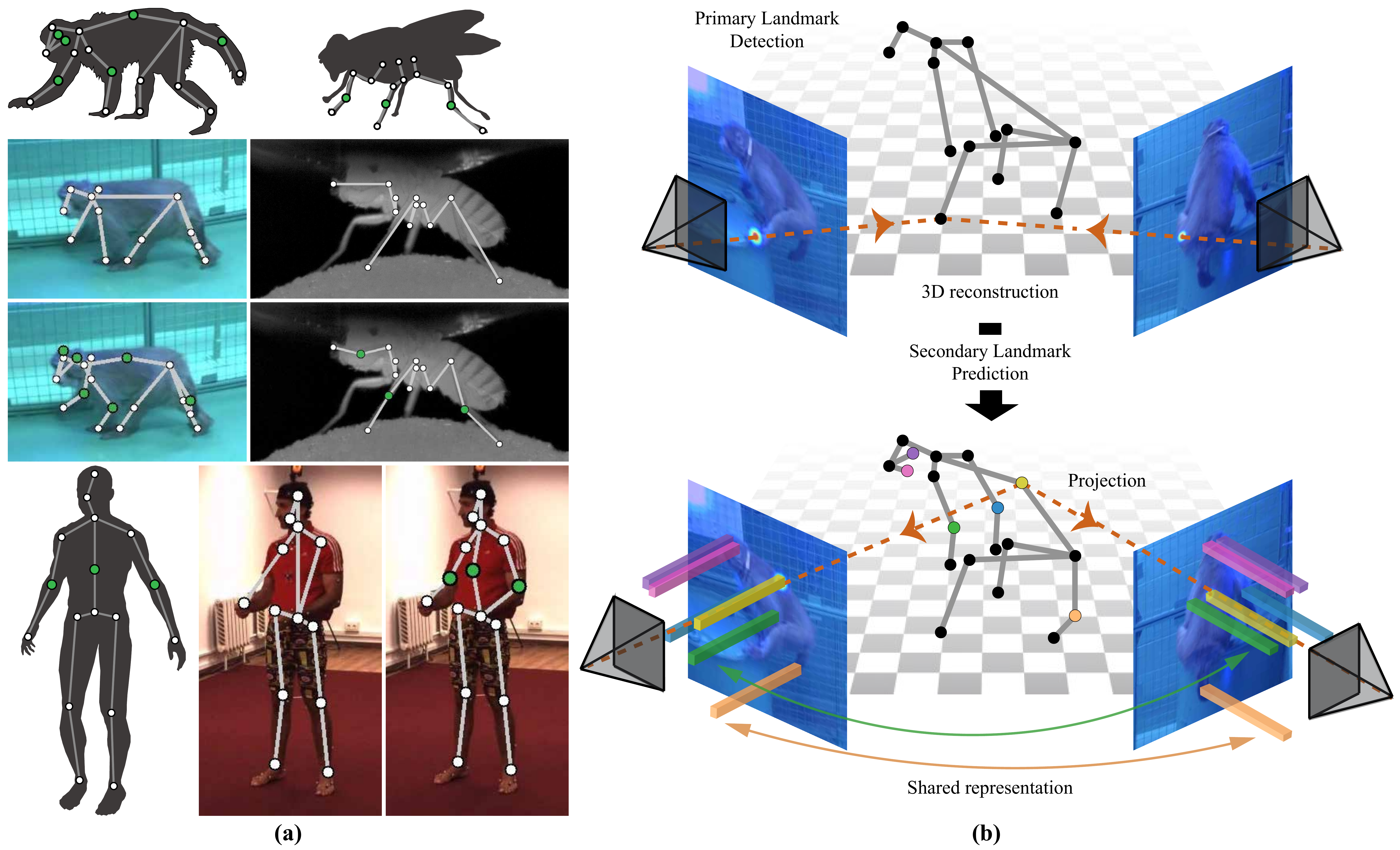} 
\end{center}
\vspace{-5mm}
\caption{This paper presents a novel semi-supervised learning approach to detect the secondary landmarks of different species, e.g., monkeys, humans and flies, by using unlabeled multiview images. (a) The primary landmarks (white circles) on body extremities characterize the overall pose  where a number of labeled data are available. On the other hand, the secondary landmarks (green circle) specify the fine-grained geometry of dynamic organisms, which is specific to each behavioral task where attaining a large labeled dataset is challenging. (b) Given multiview cameras, we propose a self-supervised learning method to estimate the secondary landmarks such as elbow joint (green). We leverage a shared representation between primary and secondary landmarks to enforce geometric consistency.}
\label{fig:teaser}
\end{figure*}

These tracking algorithms are trained by use of datasets that are annotated manually by crowd-workers or non-experts who can specify the locations of primary landmarks, that is, ones that correspond to the visually distinctive features, e.g., major body extremities such as the wrist, foot, and nose. Secondary landmarks, on the other hand, typically characterize the fine-grained geometry of the subjects, e.g., an interphalangeal joint in a toe for arthritis assessment. These secondary landmarks are visually and geometrically ambiguous; consistently annotating them often requires expert knowledge in gross animal anatomy. Further, they are often task specific, and therefore, are not included in many existing landmark datasets. For example, the OpenMonkeyPose dataset~\cite{Bala2020} does not include the elbow, tail, and ear of macaques as shown in Figure~\ref{fig:teaser}, which can be critical for studying social interactions. These issues thus present a major impediment in obtaining a large annotated secondary landmark dataset comparable to that of the primary landmarks.

We present a new method to annotate secondary landmarks in a self-supervised way by utilizing unlabeled multi-view images as shown in Figure~\ref{fig:teaser}. Our key insight is that there exists a strong spatial relationship between the primary and secondary landmarks, which implies that the primary landmarks (known) can be used to predict the secondary landmarks (unknown). This is possible because with a few exceptions, the primary landmarks on body extremities span a wide range of motion and deformation, and it is, therefore, likely to include the movement of the secondary landmarks. For instance, a secondary elbow landmark is close to the primary wrist and shoulder landmarks, which are strongly predictive of the elbow landmark. We formulate this secondary landmark prediction problem as learning a visual representation shared between the primary and secondary landmarks.

Existing image based learning approaches~\cite{gunel:2019,Bala2020} learn a visual representation in two dimensions (2D) without reasoning about underlying three dimensional (3D) geometry. 2D landmarks are a product of 3D landmarks and camera projection (3D to 2D), and therefore, learning a 2D representation alone implies learning an additional signal of camera projection. To learn the camera projection, larger annotated data seen from many viewpoints are needed. We argue that this limitation can be addressed by learning a 3D representation by factoring out the camera projection. Our 3D representation can therefore be compact, which can be learned from a substantially smaller number of annotated images. From our linear subspace analysis, we demonstrate the effectiveness of the 3D representation to express the joint subspace of the primary and secondary landmarks. 

Based on our hypothesis, we present a method to learn a coherent 3D spatial representation shared between the primary and secondary landmarks using multiview images, which allows us to self-supervise a 2D secondary landmark detector. We model this shared representation using a predictive pose model that predicts the 3D locations of the secondary landmarks given that of the primary landmarks. This 3D pose predictor is designed to be agnostic to viewpoints, which allows learning the compact representation with a small number of labeled data. The predicted 3D pose that includes the secondary landmarks is projected onto the image to supervise the secondary landmark detector (i.e., ensuring geometric consistency). Our approach differs from existing representation learning frameworks that learn the shared representation directly from 2D images~\cite{Simon:2017,yao:2019,epipolartransformers, gunel:2019} because in those frameworks, the representation needs to take into account camera projection (viewpoint), which requires a good deal of labeled data. Our approach, then, is distinguished by its very low requirements for labeled data from secondary landmarks. 

Another notable feature of our approach is that we employ multiview contrastive learning~\cite{1467314} to maximize discriminativity and/ uniqueness in the learned representation. That is, the landmarks with the same class are expected to be close in their feature space while the landmarks belonging to different classes are distant. For the secondary landmarks, we maximize the correlation between the visual features of the same landmarks while minimizing that of the different landmarks. This contrastive learning that is agnostic to the labels, in particular, plays a major role in representation learning when the number of labeled data is limited. Our approach is effective and generalizable. We show that the secondary landmarks can be reliably detected by annotating a fraction of data (using less than 10\% of data). With the learned detector, we track the secondary landmarks of organisms with diverse kinematic topologies including humans, flies, and macaques. 

%% file: 02_Result.tex
\section*{Results}
\begin{figure*}[t]
\begin{center}
\includegraphics[width=0.95\textwidth]{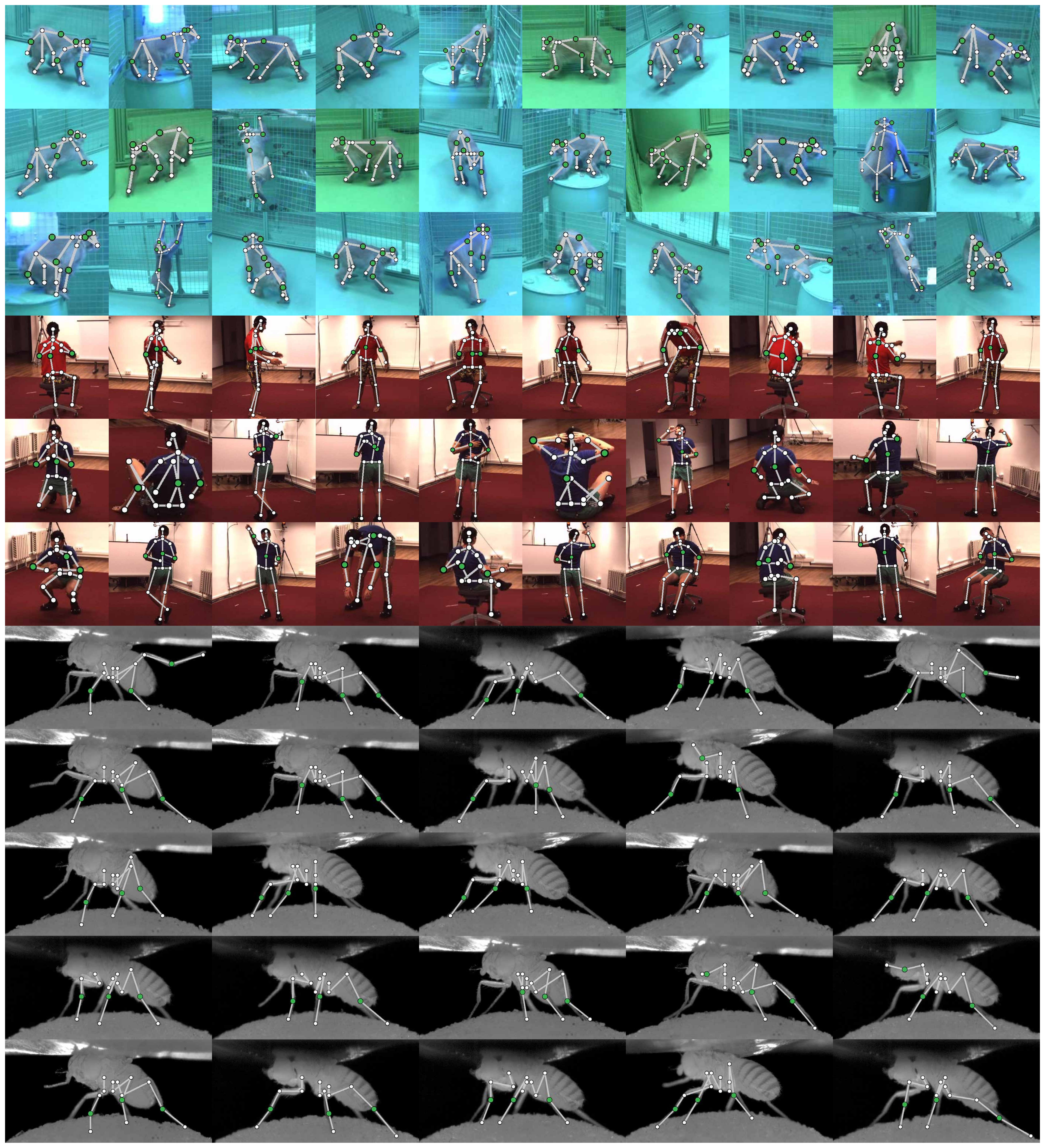} 
\end{center}
\vspace{-4mm}
\caption{Qualitative results of secondary landmark detection on OpenMonkeyPose, Human3.6M, and DeepFly3D datasets. White and green circles are the primary and secondary landmarks, respectively. For monkey dataset, we consider the elbows, ears, spine and mid-portion of tail as secondary landmarks for detection. For human dataset, we choose the elbows and spine as secondary landmarks. For flies, the three tibia-tarsus joints on the left-hand side limbs are used as secondary landmarks.}
\label{fig:qualitative}
\end{figure*}

We evaluate our method on existing real-world datasets including OpenMonkeyPose~\cite{Bala2020}, Human3.6M~\cite{IonescuSminchisescu11, h36m_pami}, and DeepFly~\cite{gunel:2019}. Figure~\ref{fig:qualitative} illustrates the qualitative results of our approach on the three datasets. For the Human3.6M and OpenMonkeyPose datasets, we observe that the elbow joints (secondary landmarks) are accurately estimated by using a strong spatial relationship between the primary and secondary landmarks. A similar observation can be made in case of the flies where the tibia-tarsus joints are accurately predicted.

\subsection*{Shared Representation Analysis}
Our method is built upon the main hypothesis that there exists a shared representation between the primary and secondary landmarks. This indicates that the motion of the secondary landmarks can be expressed by that of the primary landmarks. We validate this hypothesis using a linear subspace analysis. 

Consider a pose vector that is made of the primary and secondary landmarks $\mathbf{V} = \begin{bmatrix}\mathcal{Z}^\mathsf{T} & \mathcal{X}^\mathsf{T}\end{bmatrix}$ where $\mathcal{Z} = \begin{bmatrix}\mathbf{Z}_1^\mathsf{T}&\cdots & \mathbf{Z}_P^\mathsf{T}\end{bmatrix}^\mathsf{T}$ and $\mathcal{X} = \begin{bmatrix}\mathbf{X}_1^\mathsf{T}&\cdots & \mathbf{X}_S^\mathsf{T}\end{bmatrix}^\mathsf{T}$ are the set of the primary and secondary landmarks, respectively, i.e., $\mathbf{Z},\mathbf{X}\in\mathds{R}^3$ are the 3D coordinates of the primary and secondary landmarks. $P$ and $S$ are the number of primary and secondary landmarks, respectively. We learn a set of linear bases that span the joint space of the primary and secondary landmarks using principal coordinate analysis (PCA):
\begin{align}
\mathbf{V} = \sum_{i=1}^B \mathbf{b}_i \alpha_i + \overline{\mathbf{b}} 
\end{align}
where $B$ is the number of bases, $\mathbf{b}_i$ is the $i^{\rm th}$ linear basis, $\alpha_i$ is its coefficient, and $\overline{\mathbf{b}}$ is the mean pose. This joint space describes how the secondary landmark is related to the spatial configuration of the primary landmarks.

Given the joint space, we measure the reconstruction error of the secondary landmarks from the primary landmarks:


\begin{align}
E(\mathcal{X}) & = \|\mathcal{X} - \sum_{i=1}^B \mathbf{b}^x_i \alpha^*_i - \overline{\mathbf{b}}^x\|^2,\nonumber \\
\{\alpha^*_i\}_{i=1}^B &= \underset{\{\alpha_i\}_{i=1}^B}{\operatorname{argmin}}~\|\mathcal{Z} - \sum_{i=1}^B \mathbf{b}^z_i \alpha^*_i - \overline{\mathbf{b}}^z\|^2 &&
\label{Eq:equation2}
\end{align}

where $E(\mathcal{X})$ is the reconstruction error of the secondary landmarks $\mathcal{X}$. We decompose the basis into the primary landmark basis, $\mathbf{b}_i^z$, and the secondary landmark basis, $\mathbf{b}_i^x$, i.e., $\mathbf{b}_i = \begin{bmatrix}{\mathbf{b}_i^z}^\mathsf{T}&{\mathbf{b}_i^x}^\mathsf{T}\end{bmatrix}^\mathsf{T}$. Similarly,  $\mathbf{b}_i^x$, i.e., $\overline{\mathbf{b}} = \begin{bmatrix}{\overline{\mathbf{b}}^z}^\mathsf{T}&{\overline{\mathbf{b}}^x}^\mathsf{T}\end{bmatrix}^\mathsf{T}$. 

The secondary landmarks are reconstructed by minimizing the reconstruction error of the primary landmarks. If the primary and secondary landmarks share a joint space, minimizing the reconstruction error of the primary landmarks must minimize that of the secondary landmarks.

\begin{figure}[t]
\begin{center}
\includegraphics[width=0.48\textwidth]{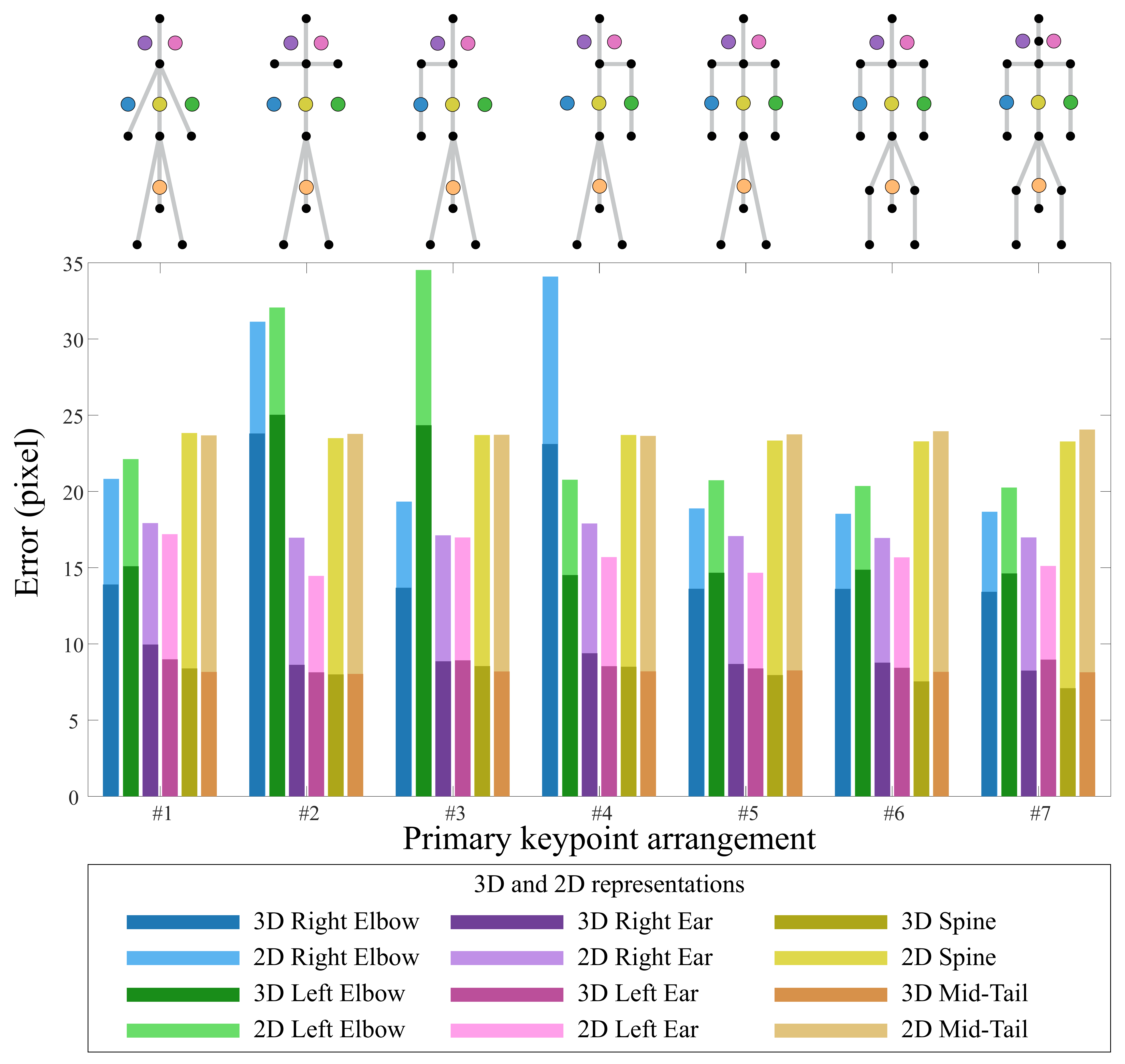} 
\end{center}
\vspace{-4mm}
\caption{Shared joint space relation between secondary and primary landmarks demonstrated on OpenMonkeyPose dataset. The lighter shades indicate the reconstruction error of secondary landmarks when predefined 2D primary landmark coordinates are used. The darker shades indicate the reconstruction error when 3D primary landmark coordinates are used. The 3D representation is highly effective to model the shared space between the primary and secondary landmarks regardless of the configuration of the primary landmarks. The reconstuction error of the 3D representation is 30\%- 60\% lower than that of the 2D representation.}
\label{fig:analysis}
\end{figure}

Figure~\ref{fig:analysis} illustrates a comparison of reconstruction error measured by Equation~(\ref{Eq:equation2}) for 2D (light bar) and 3D (dark bar) shared representations on OpenMonkeyPose dataset. We consider seven primary landmark configurations to reconstruct the secondary landmarks as shown in the top row. For all configurations, the 3D representation that factors out camera projection shows 30.2-60.6\% of error reduction compared to the 2D representation, i.e., the primary landmarks can predict the secondary landmarks more accurately through 3D representation. Among primary landmark configurations, the second configuration, i.e., the absence of wrist joints (primary) leads to an erroneous reconstruction of elbow joints (secondary). This aligns with an intuition that the motion of the secondary landmarks can be better predicted if they can be spanned by the range of motion of the primary landmarks, i.e., two joint extremities (e.g., shoulder and wrist) connecting to a secondary (e.g., elbow) are known. Similar observations can be made in the configurations \#3 and \#4, where an absence of the left wrist-shoulder and right wrist-shoulder, leads to inaccurate detection of left and right elbows respectively.

\begin{figure*}[t]
\centering
\subfloat[Reconstruction error of secondary landmarks with respect to predefined primary landmark configurations demonstrated on OpenMonkeyPose dataset. The top row indicates the primary landmark configurations. The secondary landmarks are shown in colored circles, where the size of the circles is proportional to the reconstruction error.]{\includegraphics[width=0.85\linewidth]{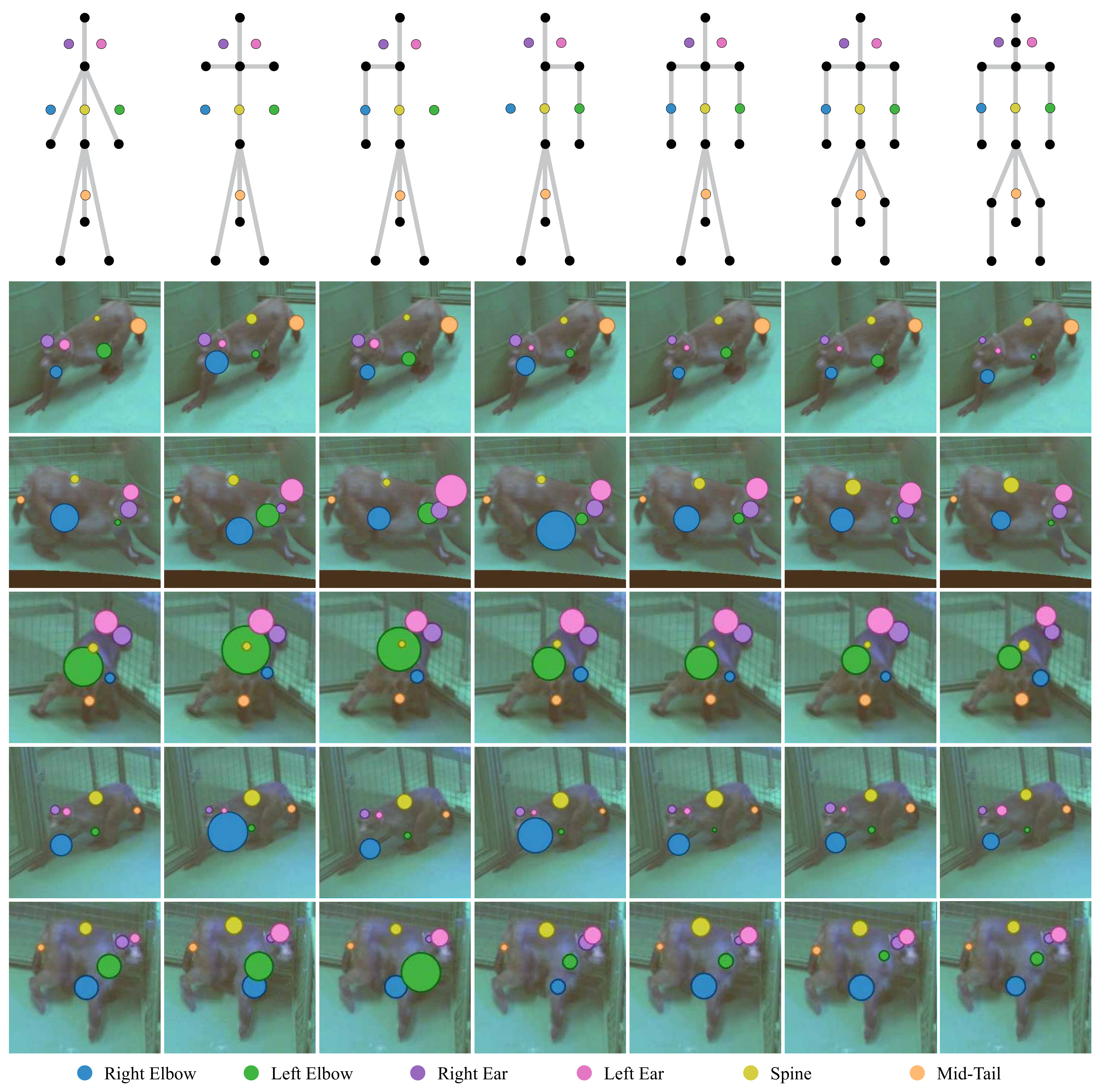}\label{fig:primary_secondary}} \\
\subfloat[PCKh for Right Elbow (Blue circles)]{\includegraphics[width=0.33\linewidth]{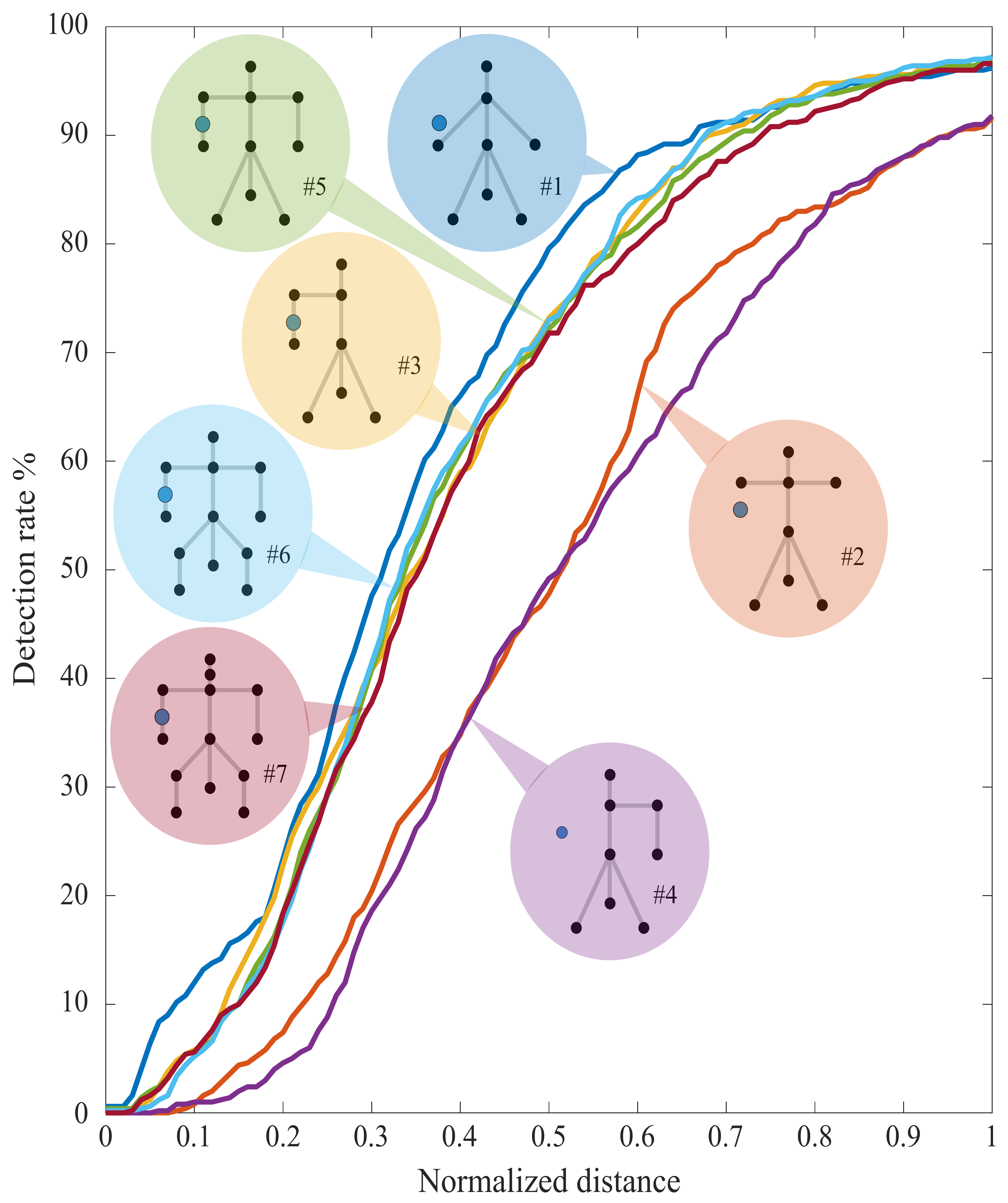}\label{fig:relbow}}
\subfloat[PCKh for Left Elbow (Green circles)]{\includegraphics[width=0.33\linewidth]{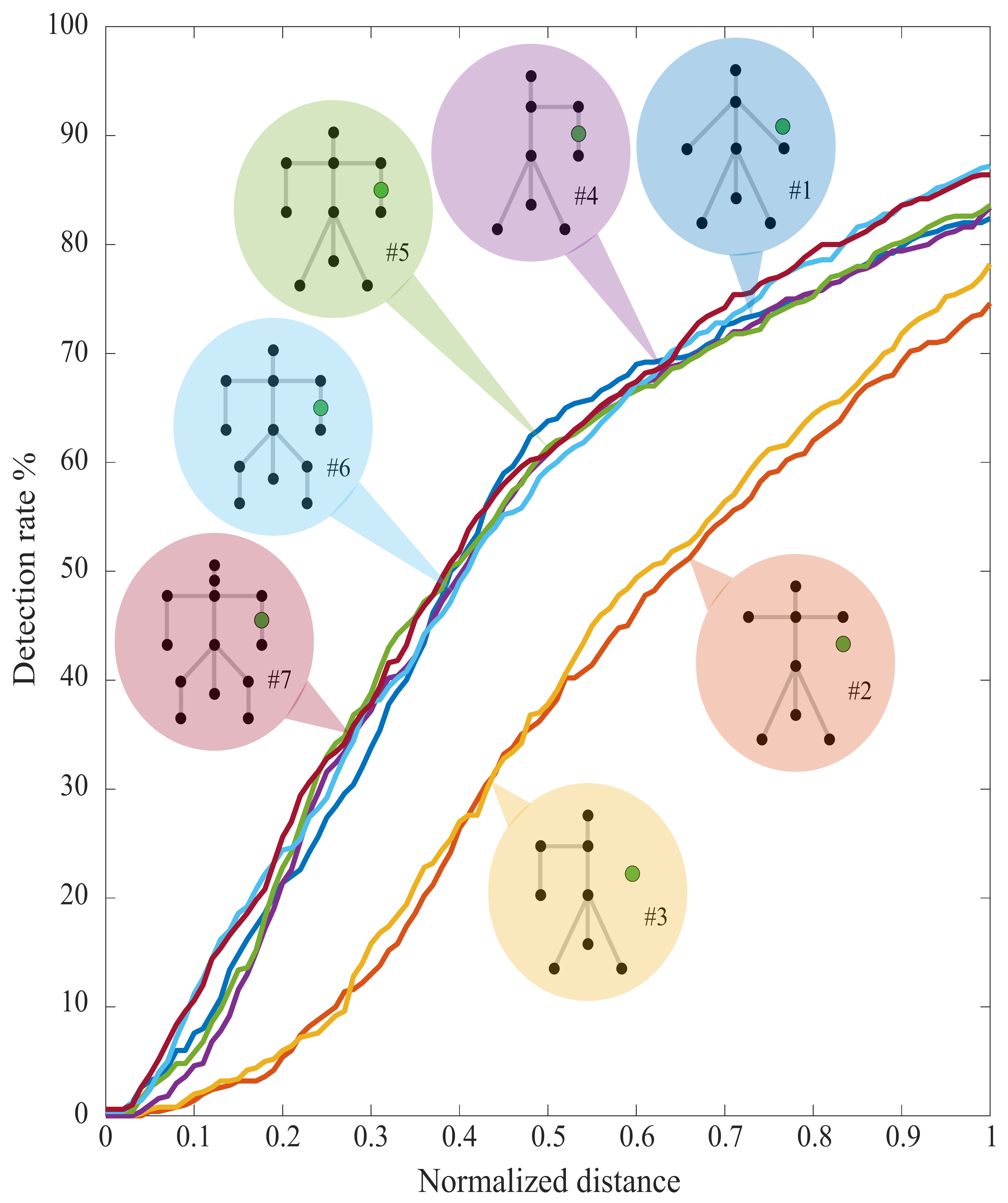}\label{fig:lelbow}}
\subfloat[PCKh for Spine (Yellow circles)]{\includegraphics[width=0.33\linewidth]{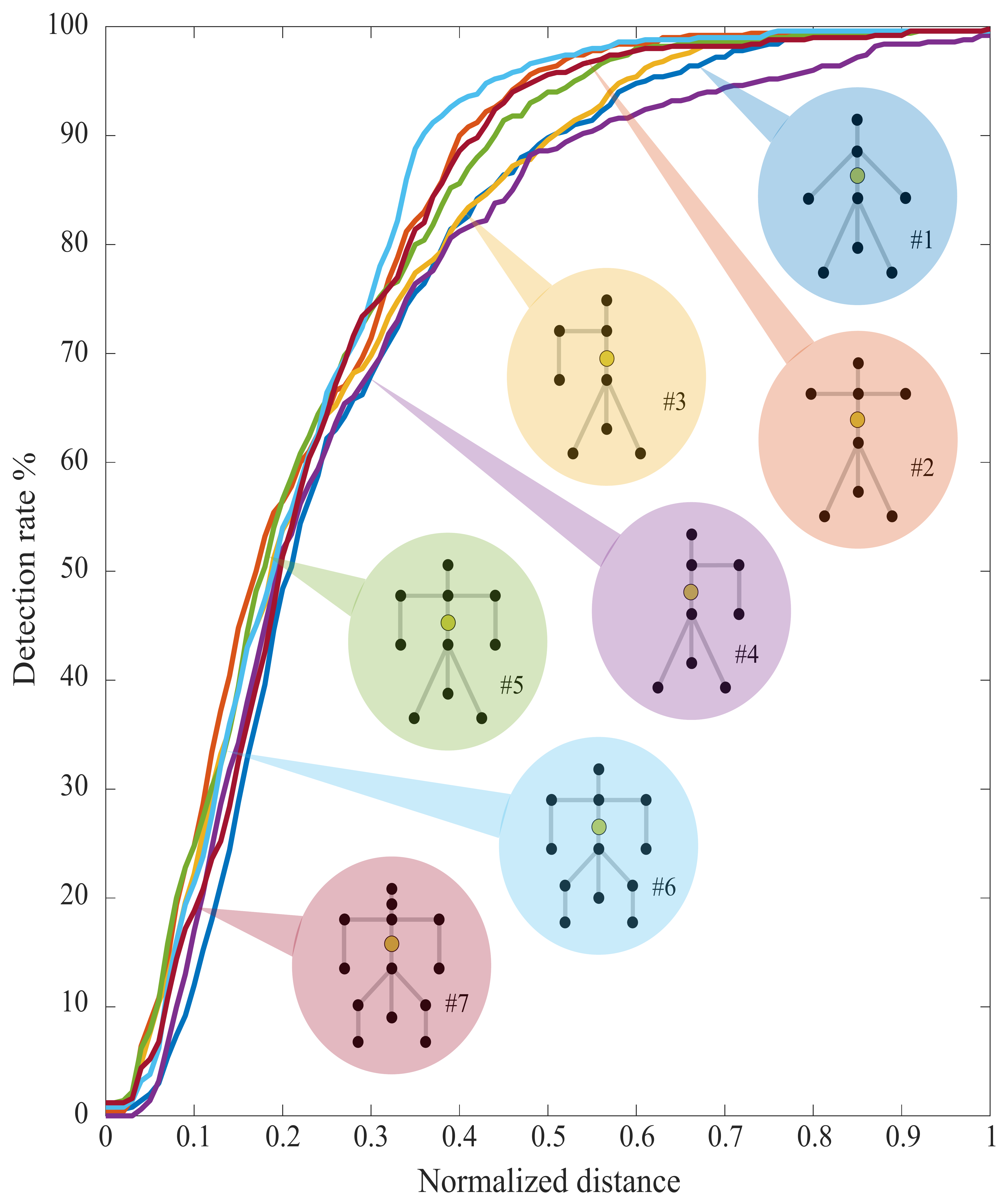}\label{fig:spine}}
\vspace{-3mm}
\caption{Reconstruction error visualization and PCKh curves for different arrangements to reconstruct the secondary landmarks}
\label{fig:primary_secondary_PCK}
\end{figure*}


Figure~\ref{fig:primary_secondary} shows the reconstruction error of the secondary landmarks with respect to the primary landmark configurations, where the size of the circles is proportional to the reconstruction error. For example, the secondary landmark reconstruction error for the left elbow and right elbow shown in the third and fourth columns are higher as compared to other secondary landmarks.



Figure~\ref{fig:relbow}, ~\ref{fig:lelbow} and~\ref{fig:spine} show the detection rate of the secondary landmarks using probability of correct keypoint metric~\cite{andriluka14cvpr} with an error tolerance of the size of head (PCKh). For instance, in the configuration \#2 and \#4, the lack of the shoulder or wrist primary landmarks lead to inaccurate reconstruction of the right elbow secondary landmark (Figure~\ref{fig:lelbow}).  This indicates that spatial adjacency plays a pivotal role in determining the location of a given landmark, in this case the right elbow. Similar observations can be made in Figure~\ref{fig:spine}.

\subsection*{Impact of Contrastive Learning}
In addition to joint representation learning, we leverage unsupervised contrastive feature learning to learn a discriminative representation. We maximize the feature correlation between the same landmarks while minimizing it between different landmarks.

\begin{figure*}[t]
\centering  
\vspace{-3mm}
\captionsetup[subfigure]{labelformat=empty}
\subfloat{\label{fig:umap}\includegraphics[height=0.48\textwidth]{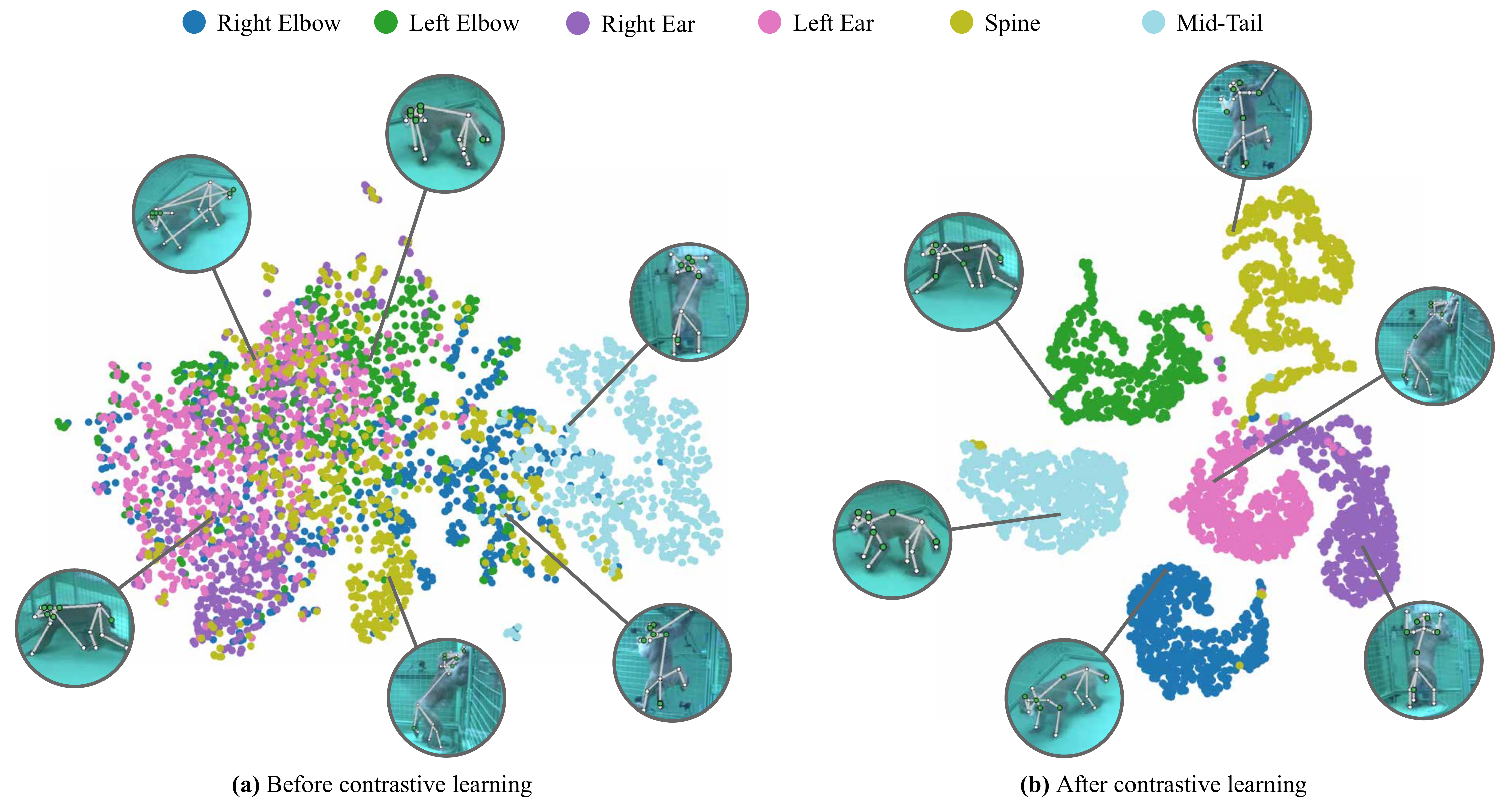}} \\
\subfloat[\textbf{(c)} Before contrastive learning]{\label{Fig:supp1a}\includegraphics[height=0.35\textwidth]{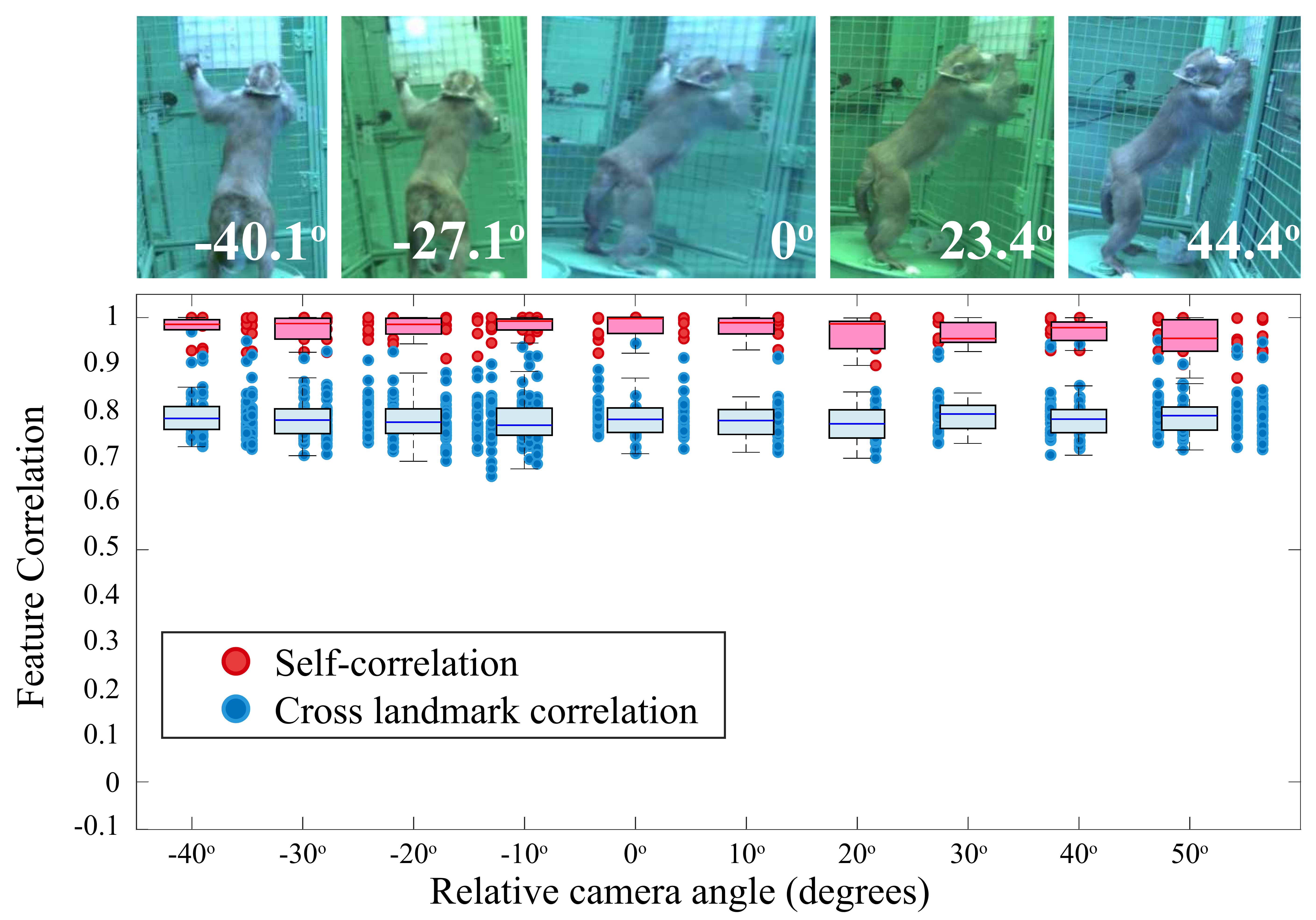}}~~
\subfloat[\textbf{(d)} After contrastive learning]{\label{Fig:supp1b}\includegraphics[height=0.35\textwidth]{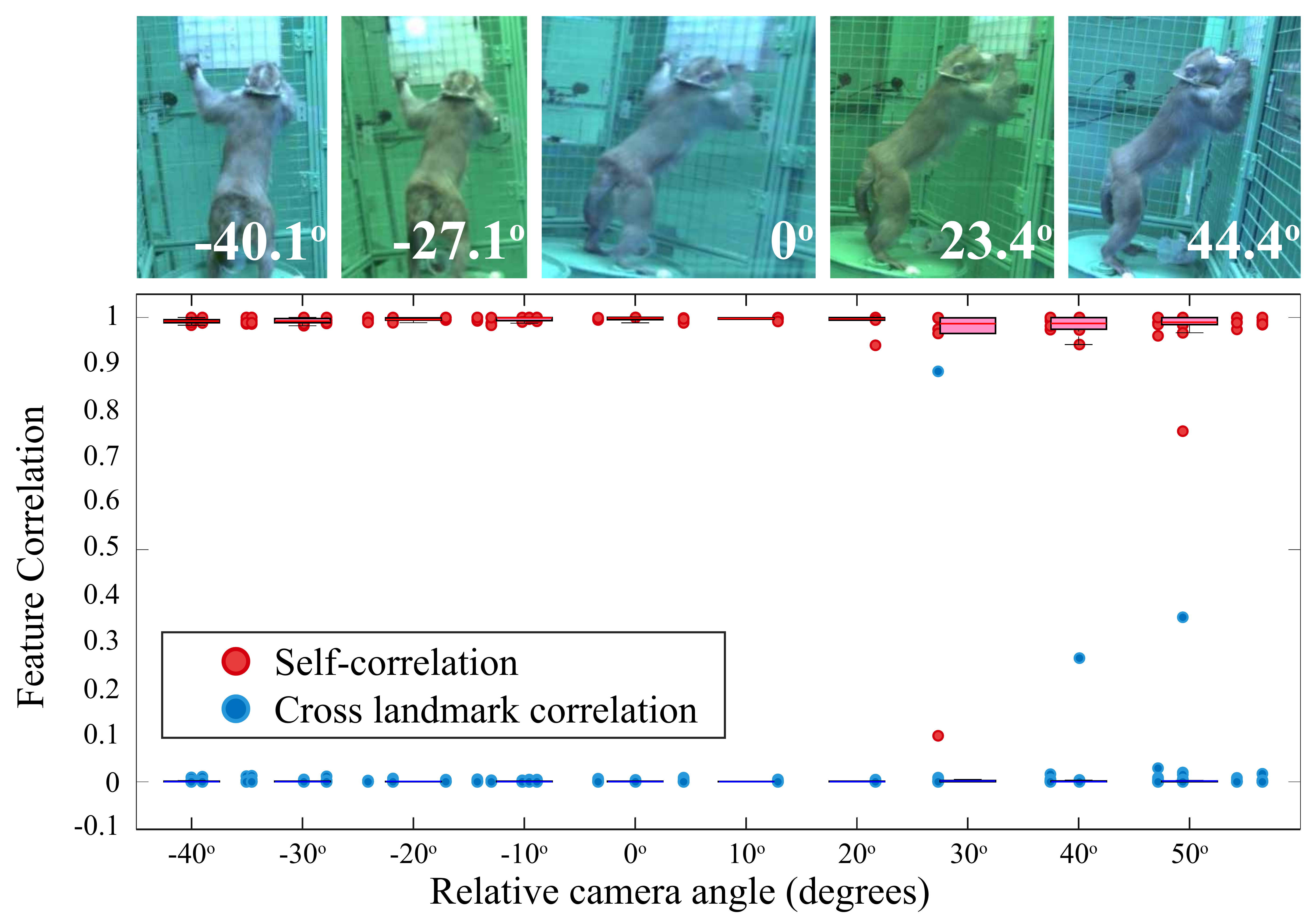}}
\vspace{-3mm}
\caption{Contrastive learning enforces maximizing feature correlation between the same landmark across views and minimizing that between different landmarks. (b) With contrastive learning the red points (self-landmark correlation) are far apart from the blue points (cross-landmark corrleation). } 
\label{fig:supp1}
\end{figure*}

Figure~\ref{fig:umap} and~\ref{Fig:supp1a} illustrate the visualization of UMAP of landmark feature space with and without contrastive learning. Unlike the feature distribution without contrastive learning, that with contrastive learning aligns the features of the same secondary landmarks, forming distinctive clusters with respect to the secondary landmarks regardless of views and poses.

Figure~\ref{Fig:supp1b} and \textcolor{blue}{5d} illustrates the feature correlation of the landmarks across views. The red points indicate the correlation among the same secondary landmarks from different views, denoted as self-correlation. The blue points indicate the correlation among different secondary landmarks, denoted here as cross landmark correlation. The one with contrastive learning produces the visual feature of a secondary landmark that is highly correlated with that of another secondary landmark (most correlation is higher than 0.7). In contrast, this correlation is minimized with contrastive learning, which makes the visual feature more discriminative.

\paragraph{Ablation Study} We conduct an ablation study to measure the impact of each component. (1) $\mathcal{L}_{\rm L}$: a landmark detection model that is fully supervised by the labeled data. This model is equivalent to the supervised learning models. (2) $\mathcal{L}_{\rm L}+\mathcal{L}^t_{\rm U}$: a semi-supervised learning model that uses the reprojection of triangulated secondary landmarks to self-supervise the secondary landmark locations. This method is an application of Günel et al.~\cite{gunel:2019} for the task of secondary landmark prediction. (3) $\mathcal{L}_{\rm L}+\mathcal{L}^g_{\rm U}$: a semi-supervised learning model that uses 3D shared representation to self-supervise the secondary landmark locations. 3D reconstructed secondary landmarks are projected onto multiview images to supervise the landmark detector (minimizing reprojection error). (4) $\mathcal{L}_{\rm L}+\mathcal{L}^g_{\rm U}+\mathcal{L}^c_{\rm U}$: a semi-supervised learning model (our full model) that use both the shared representation and contrastive learning for self-supervision. It minimizes geometric error but also maximizes visual feature correlation, resulting in a view-invariant representation.

Table~\ref{tab:1} summarizes the performance of each method measured at PCKh@$t$ where $t$=\{0.25, 0.5, 0.75\} on the secondary landmarks for three datasets. In general, semi-supervised learning $\mathcal{L}^g_{\rm U}$ that uses multiview unlabeled data significantly outperforms the supervised learning method $\mathcal{L}_{\rm L}$. Further, the model trained with contrastive learning,  $\mathcal{L}_{\rm L}+\mathcal{L}^g_{\rm U}+\mathcal{L}^c_{\rm U}$, improves the performance on generalization. In particular, the left-ear, right-ear, spine, and mid-tail joints in OpenMonkeyPose, right-elbow, left-elbow and spine joints in Humans 3.6M and J3, J8 and J13 joints in DeepFly3D show significant improvement. 

\begin{table*}[t]
\hspace{-4mm}
\footnotesize
\begin{tabular}{p{1.8cm}p{0.5cm}cccccc|cccc|cccc|c}
\toprule 
\multicolumn{2}{c}{} & \multicolumn{7}{c}{OpenMonkeyPose~\cite{Bala2020}}& \multicolumn{4}{c}{Human3.6M~\cite{IonescuSminchisescu11, h36m_pami}} & \multicolumn{4}{c}{DeepFly~\cite{gunel:2019}}\\
\cmidrule(lr){3-9}\cmidrule(lr){10-13}\cmidrule(lr){14-17}
Methods
& $t$ & R.Elb & L.Elb & R.Ear & L.Ear & Spine & M.Tail & Mean & R.Elb & L.Elb & Spine & Mean & J3 & J8 & J13 & Mean\\
\cmidrule(lr){1-2}\cmidrule(lr){3-9}\cmidrule(lr){10-13}\cmidrule(lr){14-17}
$\mathcal{L}_{\rm L}$ & 0.25 & \textbf{0.31} & 0.21 & 0.49 & 0.52 & 0.41 & 0.34 & 0.37& 0.07 & 0.04 & 0.04 & 0.05 & 0.18 & 0.21 & 0.11 & 0.17\\
$\mathcal{L}_{\rm L} + \mathcal{L}_{\rm U}^t$ & 0.25 & 0.24 & 0.25 & \textbf{0.50} & 0.52 & 0.41 & 0.35 & 0.37& 0.02 & 0.03 & 0.04 & 0.03 & 0.04 & 0.30 & 0.20 & 0.18\\
$\mathcal{L}_{\rm L} + \mathcal{L}_{\rm U}^g$ & 0.25 & 0.24 & 0.20 & 0.49 & 0.50 & 0.40 & \textbf{0.39} & 0.37& 0.07 & 0.11 & \textbf{0.30} & 0.16 & 0.11 & 0.34 & 0.27 & 0.24\\
$\mathcal{L}_{\rm L} + \mathcal{L}_{\rm U}^g+ \mathcal{L}_{\rm U}^c$& 0.25 & 0.27 & \textbf{0.25} & 0.49 & \textbf{0.52} & \textbf{0.41} & 0.36 & \textbf{0.38} & \textbf{0.23} & \textbf{0.21} & 0.27 & \textbf{0.24} & \textbf{0.52} & \textbf{0.58} & \textbf{0.39} & \textbf{0.50}\\ 

\cmidrule(lr){1-2}\cmidrule(lr){3-9}\cmidrule(lr){10-13}\cmidrule(lr){14-17}
$\mathcal{L}_{\rm L}$  & 0.50 & 0.63 & 0.47 & 0.85 & \textbf{0.89} & 0.74 & 0.72 & 0.72 & 0.16 & 0.10 & 0.11 & 0.12 & 0.28 & 0.38 & 0.26 & 0.31\\
$\mathcal{L}_{\rm L} + \mathcal{L}_{\rm U}^t$  & 0.50 & 0.60 & 0.51 & 0.86 & 0.87 & 0.80 & 0.74 & 0.73& 0.04 & 0.09 & 0.14 & 0.09 & 0.07 & 0.41 & 0.35 & 0.28\\
$\mathcal{L}_{\rm L} + \mathcal{L}_{\rm U}^g$  & 0.50 & 0.62 & 0.54 & 0.86 & 0.86 & 0.80 & \textbf{0.79} & 0.74& 0.25 & 0.38 & 0.65 & 0.43 & 0.19 & 0.58 & 0.50 & 0.42\\
$\mathcal{L}_{\rm L} + \mathcal{L}_{\rm U}^g+ \mathcal{L}_{\rm U}^c$ & 0.50 & \textbf{0.63} & \textbf{0.54} & \textbf{0.91} & 0.88 & \textbf{0.80} & 0.74 & \textbf{0.75}& \textbf{0.58} & \textbf{0.47} & \textbf{0.67} & \textbf{0.57} & \textbf{0.83} & \textbf{0.76} & \textbf{0.79} & \textbf{0.79}\\ 

\cmidrule(lr){1-2}\cmidrule(lr){3-9}\cmidrule(lr){10-13}\cmidrule(lr){14-17}
$\mathcal{L}_{\rm L}$  & 0.75 & 0.73 & 0.65 & 0.95 & \textbf{0.97} & 0.83 & 0.90 & 0.84& 0.25 & 0.16 & 0.21 & 0.21 & 0.33 & 0.44 & 0.34 & 0.37\\
$\mathcal{L}_{\rm L}+ \mathcal{L}_{\rm U}^t$ & 0.75 & 0.77 & 0.70 & 0.93 & 0.95 & 0.92 & 0.87 & 0.85 & 0.08 & 0.14 & 0.22 & 0.15 & 0.08 & 0.49 & 0.39 & 0.32\\
$\mathcal{L}_{\rm L}+ \mathcal{L}_{\rm U}^g$ & 0.75 & \textbf{0.78} & 0.72 & 0.93 & 0.95 & 0.94 & 0.91 & 0.86 & 0.45 & 0.59 & 0.84 & 0.64 & 0.23 & 0.63 & 0.56 & 0.47\\
$\mathcal{L}_{\rm L} + \mathcal{L}_{\rm U}^g+ \mathcal{L}_{\rm U}^c$ & 0.75 & 0.77 & \textbf{0.75} & \textbf{0.95} & 0.96 & \textbf{0.95} & \textbf{0.91} & \textbf{0.88} & \textbf{0.76} & \textbf{0.64} & \textbf{0.85} & \textbf{0.75} & \textbf{0.91} & \textbf{0.82} & \textbf{0.96} & \textbf{0.90}\\ \bottomrule 

\end{tabular}
\vspace{-3mm}
\caption{We conduct an ablation study to evaluate the impact of each component on secondary landmark detection. PCKh@$t$ is used as the evaluation metric. (1) $\mathcal{L}_{\rm L}$ is standard supervised learning using both the labeled primary and secondary landmarks. (2) $\mathcal{L}_{\rm L}+\mathcal{L}^t_{\rm U}$ is a semi-supervised learning model that uses the reprojection of triangulated secondary landmarks to self-supervise the secondary landmark locations. (3) $\mathcal{L}_{\rm L}+\mathcal{L}_{\rm U}^g$ is semi-supervised learning where we minimize the geometric error for multiview supervision. (4) $\mathcal{L}_{\rm L} + \mathcal{L}_{\rm U}^g+ \mathcal{L}_{\rm U}^c$ is our full model that enforces both shared representation and contrastive learning to learn the secondary landmarks. In general, the semi-supervised learning approach outperforms the supervised approach. }
\label{tab:1}
\end{table*}

\begin{figure*}[t]
\centering
\vspace{-3mm}
\subfloat[OpenMonkeyPose]{\includegraphics[width=0.33\linewidth]{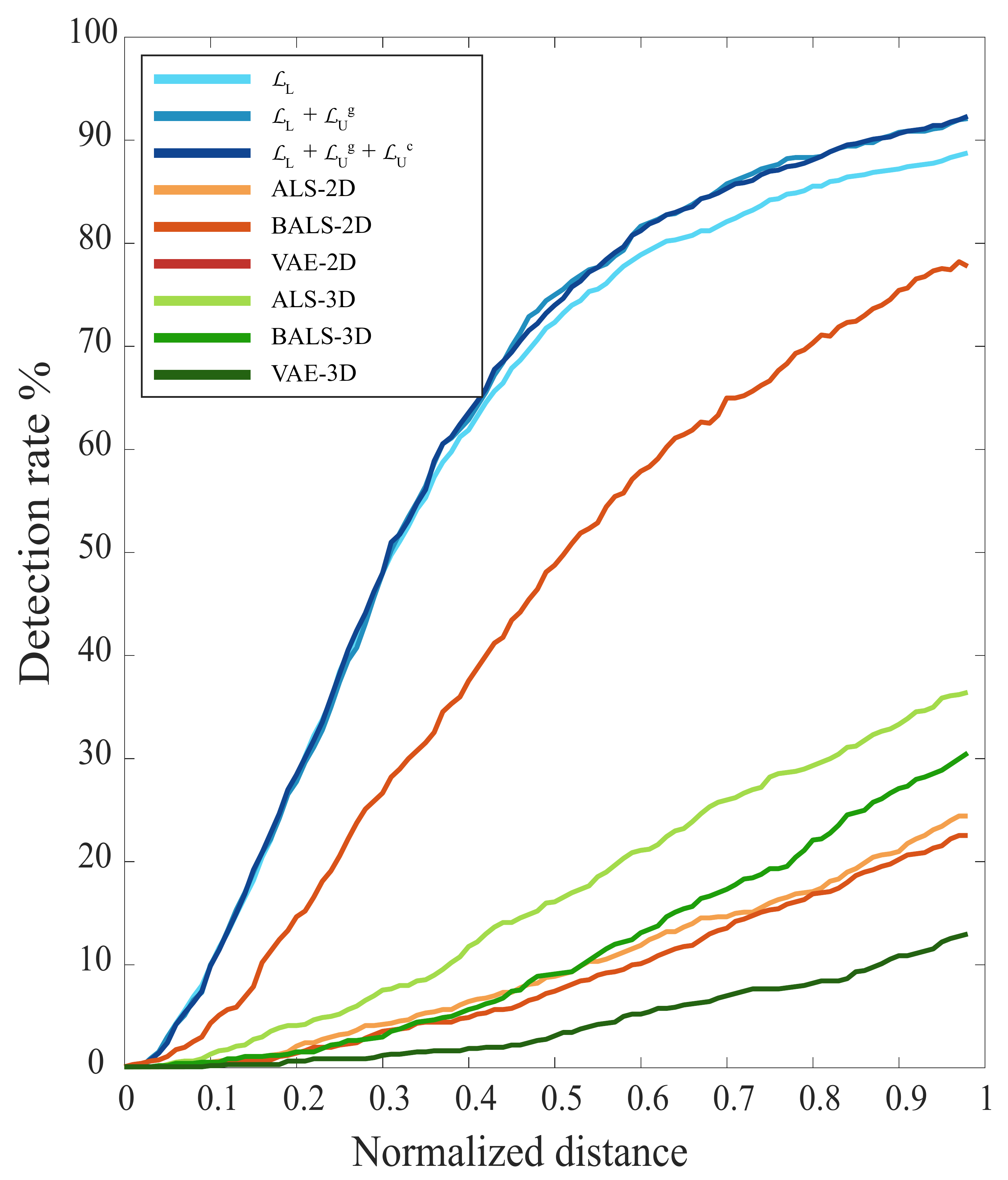}\label{fig:monkey}}
\subfloat[Human3.6M]{\includegraphics[width=0.33\linewidth]{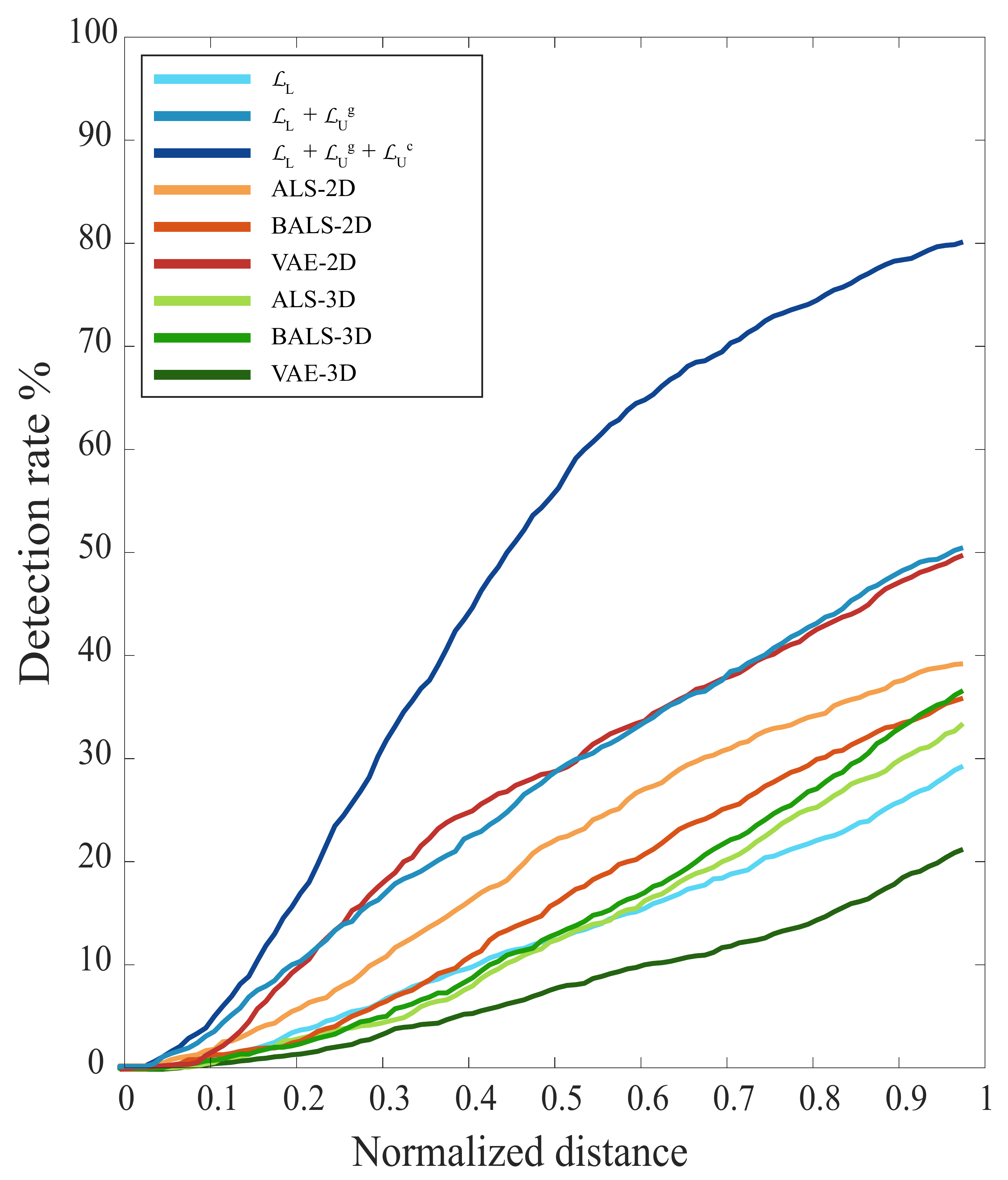}\label{fig:human}}
\subfloat[DeepFly]{\includegraphics[width=0.33\linewidth]{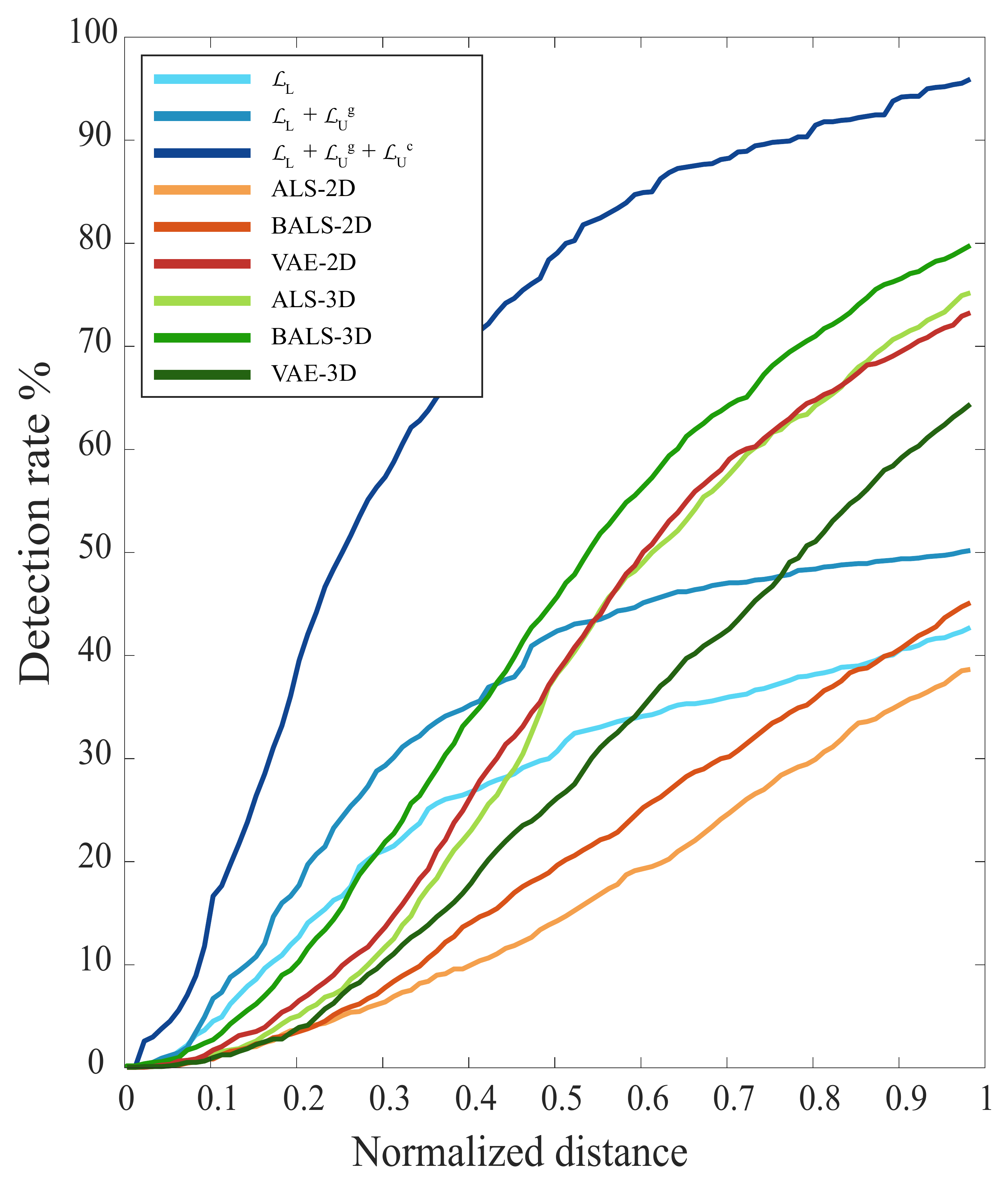}\label{fig:flies}}
\vspace{-3mm}
\caption{PCKh curves for (a) monkeys, (b) humans and (c) flies. The proposed method outperforms the listed baseline algorithms}
\label{Fig:supp2}
\end{figure*}

\begin{table*}[t]
\hspace{-2mm}
\footnotesize
\begin{tabular}{p{2.6cm}cccccc|cccc|cccc|c}\toprule 
\multicolumn{1}{c}{} & \multicolumn{7}{c}{OpenMonkeyPose~\cite{Bala2020}}& \multicolumn{4}{c}{Human3.6M~\cite{IonescuSminchisescu11, h36m_pami}} & \multicolumn{4}{c}{DeepFly~\cite{gunel:2019}}\\
\cmidrule(lr){2-8}\cmidrule(lr){9-12}\cmidrule(lr){13-16}
Methods & R.Elb & L.Elb & R.Ear & L.Ear & Spine & M.Tail & Mean & R.Elb & L.Elb & Spine & Mean & J3 & J8 & J13 & Mean\\ 
\midrule
ALS (2D)~\cite{matrix_factorization} & 0.02 & 0.02 & 0.09 & 0.09 & 0.21 & 0.06 & 0.09 & 0.05 & 0.05 & 0.58 & 0.23 & 0.13 & 0.25 & 0.09 & 0.16\\
BALS (2D)~\cite{svd_decomp} & 0.05 & 0.01 & 0.03 & 0.01 & 0.11 & 0.09 & 0.07 & 0.04 & 0.05 & 0.44 & 0.18 & 0.20 & 0.27 & 0.07 & 0.18\\
VAE (2D)~\cite{gaps} & 0.18 & 0.26 & 0.66 & 0.59 & 0.67 & 0.57 & 0.49 & 0.13 & 0.15 & 0.61 & 0.29 & 0.40 & 0.42 & 0.37 & 0.40\\
ALS (3D)~\cite{matrix_factorization} & 0.02 & 0.03 & 0.30 & 0.39 & 0.18 & 0.05 & 0.16 & 0.08 & 0.06 & 0.23 & 0.12 & 0.36 & 0.49 & 0.30 & 0.38\\
BALS (3D)~\cite{svd_decomp} & 0.05 & 0.08 & 0.07 & 0.15 & 0.12 & 0.07 & 0.07 & 0.11 & 0.05 & 0.23 & 0.13 & 0.56 & 0.50 & 0.31 & 0.46\\
VAE (3D)~\cite{gaps} & 0.03 & 0.02 & 0.02 & 0.03 & 0.06 & 0.02 & 0.03 & 0.19 & 0.11 & 0.55 &  0.28 & 0.21 & 0.29 & 0.29 & 0.26\\
$\mathcal{L}_{\rm L}$ & \textbf{0.63} & 0.47 & 0.85 & \textbf{0.89} & 0.74 & 0.72 & 0.72 & 0.16 & 0.10 & 0.11 & 0.12 & 0.28 & 0.38 & 0.26 & 0.31\\
$\mathcal{L}_{\rm L} + \mathcal{L}_{\rm U}^g+ \mathcal{L}_{\rm U}^c$ (Ours)  & 0.63 & \textbf{0.54} & \textbf{0.91} & 0.88 & \textbf{0.80} & \textbf{0.74} & \textbf{0.75} & \textbf{0.58} & \textbf{0.47} & \textbf{0.67} & \textbf{0.57} & \textbf{0.83} & \textbf{0.76} & \textbf{0.79} & \textbf{0.79}\\ \bottomrule 
\end{tabular}
\vspace{-3mm}
\caption{We evaluate the performance of our method on three datasets by comparing with that of the baseline approaches. We report the PCKh@0.5 values for each of the evaluated methods.}
\label{tab:2}
\end{table*}

Figure~\ref{Fig:supp2} shows the PCKh performance of the baseline methods to detect secondary landmarks on human, monkey and fly subjects. Our approach significantly outperforms the listed baseline approaches.

\subsection*{Comparison with State-of-the-Art approaches}
We compare the accuracy of the secondary landmark detection with 7 baseline algorithms. (1) Alternating least squares~\cite{matrix_factorization} (ALS): this is a matrix completion method that can predict the secondary landmarks by considering them as missing entries in a matrix and by applying rank minimization. We construct the matrix made of the coordinates of the primary and secondary landmarks. Given a set of primary landmarks, the algorithm finds a nearest neighbor from the labeled set, and completes the missing secondary landmarks by minimizing the rank of the matrix. (2) Biased alternating least squares~\cite{svd_decomp} (BALS): this is a variant of alternating least squares with weighted-$\lambda$-regularization. (3) Variational autoencoder~\cite{gaps} (VAE): this learns a latent code that can express the data distribution in the presence of missing data. A VAE has been used in 2D human pose estimation with occlusion~\cite{gaps}. We apply these three methods on the 2D and 3D secondary landmark prediction. (4) Supervised approach $\mathcal{L}_{\rm L}$: we use the labeled secondary and primary landmarks to train the landmark detector in a fully supervised manner.

We use a metric called PCKh~\cite{andriluka14cvpr} to measure accuracy of secondary landmark detection. A predicted landmark is considered as correct if it is within $tL$ pixels from the ground truth landmark, where $t$ is an error tolerance proportion given the reference length $L$: the length between the head and neck joints for the OpenMonkeyPose and Human3.6M datasets, and the length between joints {J0} and {J4} for $L$ for the DeepFly3D dataset.

We evaluate the approaches based on the accuracy of the predictions and use PCKh@0.5 as the evaluation metric summarized in Table~\ref{tab:2}. It is worth noting that while a few approaches exhibit strength for particular landmarks, none of these baseline methods consistently dominate over all the datasets. As expected, due to the limited amount of data, the supervised method performs poorly.  The quality of our secondary landmark detection strongly outperforms the baselines with a large margin.

\begin{figure*}[t]
\begin{center}
\includegraphics[width=1\textwidth]{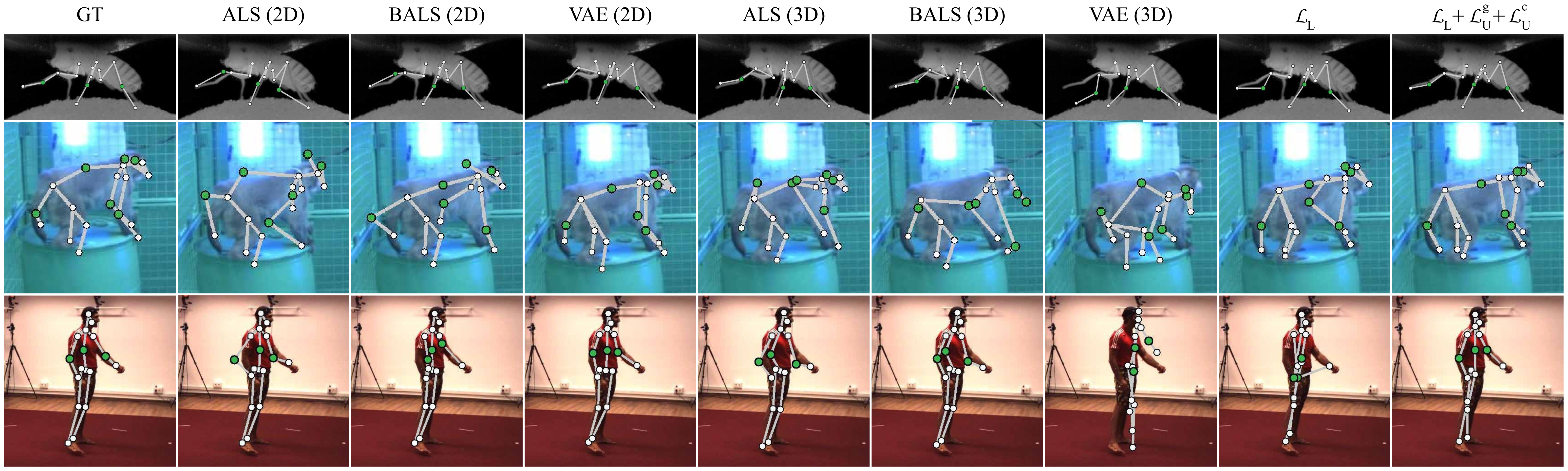} 
\end{center}
\vspace{-4mm}
  \caption{We qualitatively compare the performance of our method to detect secondary landmarks with 7 baseline methods mentioned in the main paper on DeepFly, Human3.6M, and OpenMonkeyPose datasets}
\label{fig:supp3}
\end{figure*}

In Figure~\ref{fig:supp3}, we report the qualitative comparison that shows the performance of the baselines listed in Table~\ref{tab:2}, to detect the secondary landmarks.

\begin{table}\centering
\footnotesize
\begin{tabular}{lccccc}\toprule
Method & $|\mathcal{D}_{\mathcal{X}}|/|\mathcal{D}^U_\mathcal{X}|$ & Secondary & Primary & Mean\\
\midrule
$\mathcal{L}_{\rm L}$ & 2k/140k & 0.567 & 0.693 & 0.653\\
$\mathcal{L}_{\rm L} + \mathcal{L}_{\rm U}^g+ \mathcal{L}_{\rm U}^c$   & 2k/140k & 0.413 & 0.679 & 0.595\\ \midrule

$\mathcal{L}_{\rm L}$ & 6k/140k & 0.687 & 0.701 & 0.696\\
$\mathcal{L}_{\rm L} + \mathcal{L}_{\rm U}^g+ \mathcal{L}_{\rm U}^c$ & 6k/140k & 0.662 & 0.669 & 0.667\\\midrule

$\mathcal{L}_{\rm L}$ & 10k/140k & 0.707 & 0.707 & 0.707\\
$\mathcal{L}_{\rm L} + \mathcal{L}_{\rm U}^g+ \mathcal{L}_{\rm U}^c$ & 10k/140k & 0.702 & 0.703 & 0.703\\ \midrule

$\mathcal{L}_{\rm L}$ & 14k/140k & 0.722 & 0.693 & 0.706\\
$\mathcal{L}_{\rm L} + \mathcal{L}_{\rm U}^g+ \mathcal{L}_{\rm U}^c$ & 14k/140k & \textbf{0.753} & \textbf{0.693} & \textbf{0.712}\\ \bottomrule
\end{tabular}
\vspace{-3mm}
\caption{We study the performance of the secondary landmark detection as varying the amount of labeled data $|\mathcal{D}_{\mathcal{X}}|$ while that of unlabeled multiview data $|\mathcal{D}^U_\mathcal{X}|$ remains constant.}
\label{tab:3}
\end{table}

In Table~\ref{tab:3}, we evaluate the performance of the secondary landmark detection on the OpenMonkeyPose dataset by varying the amount of labeled data, keeping the unlabeled multiview data  constant. We observe that the performance of our method increases as the amount of secondary landmark labels increases. At an observed ratio of 1:10, the proposed method outperforms the fully-supervised approach.

%% file: 04_Discussion.tex
\section*{Discussion}
We propose a new solution to a relatively under-studied problem in the field of automated behavioral tracking, that of secondary landmark detection. Unlike primary landmarks that describe generic and coarse body geometry, secondary landmarks are of particular interest because their spatial configuration specifies the fine-grained geometry of organisms. Indeed, they are particularly likely to be useful in customized behavioral tasks or to answer bespoke tracking questions. Our secondary landmark detector is learned from unlabeled multiview images in conjunction with a small set of annotated secondary landmarks. It leverages the key insight that there exists a strong spatial relationship between the primary and secondary landmarks, which allows us to learn their shared representation from unlabeled data. We develop a self-supervised predictive model that can estimate the secondary landmarks from the primary landmarks.

The spatial relationship between the primary and secondary is more apparent in 3D, i.e., it is easier to predict 3D landmarks than 2D landmarks that are a function of camera projection. Therefore, the use of multiview images is a critical element of our method. By using multiview images, we can triangulate the primary landmarks that can in turn be used to predict 3D secondary landmarks. These reconstructed 3D secondary landmarks are, in turn, projected onto each view to supervise the 2D secondary landmark detector. This process is helped by a contrastive learning scheme that learns the distinctive and unique visual representation of the secondary landmarks. These later processes are completely label agnostic, i.e., the learning relies on self-supervision. Our method is generic, applicable to diverse species, camera poses, and primary/secondary landmark configurations. Indeed, we demonstrate that our method can reliably augment landmarks with a smaller number of secondary landmarks on humans, monkeys, and flies.

Through a linear subspace analysis, we demonstrate that there exists a subspace shared between the primary and secondary landmarks, and that, this subspace can be used to predict the secondary landmarks given the primary landmarks. Especially when constrained to a limited amount of data, the 3D shared representation is more effective and expressive than the 2D shared representation, which agrees with our central hypothesis. Our 3D shared representation learning differentiates it from existing approaches~\cite{gunel:2019} that enforce learning a 2D spatial configuration without reasoning about 3D geometry. In contrast, we explicitly learn the 2D spatial configuration through the 3D shared representation that provides self-supervision to the 2D secondary landmark detector. Based on this linear analysis, we characterize the secondary landmark prediction as a function of the primary landmark configuration so that a secondary landmark can be accurately predicted.

Our secondary landmark detection can be thought of as a new landmark annotation paradigm parallel to the existing transfer learning paradigm used in neuroscience and biology. For example, DeepLabCut~\cite{Mathis2018,Mathis2019} uses a small number of annotations to learn a generalizable visual representation via transfer learning: transferring a generic image representation learned from a large image dataset such as ImageNet to the target animal images. On the other hand, our approach takes an incremental bootstrapping that can substantially reduce manual annotation efforts updating a visual representation by introducing a new set of landmarks at each time given the previously learned representation for the existing landmarks (e.g., primary landmarks). Similar to transfer learning, this bootstrapping leverages a strong prior of the visual representation, which allows building a generalizable augmented landmark detector. This, however, implies the limitation of our approach: it needs an initial good representation to start the bootstrapping process, which requires a sufficient amount of annotated data for the primary landmarks. We assume that such annotated data can be attainable from existing annotation tools such as DeepLabCut~\cite{Mathis2018,Mathis2019} or OpenMonkeyStudio~\cite{Bala2020}. 

\begin{acknowledgements}
This work was supported by
NSF IIS 2024581 to HSP, JZ, and BYH \\
NSF IIS 1846031 to HSP, PCB \\
NIMH 125377 to BYH
\end{acknowledgements}

%% file: 03_Method.tex
\section*{Method}
We present a new method to learn a secondary landmark detector given the primary landmarks and unlabeled multiview images. The \textit{primary landmarks} refer to the base landmarks that characterize the overall pose such as the body extremities, and the \textit{secondary landmarks} refer to the customized landmarks that describe their fine-grained geometry. We denote the sets of the primary and secondary landmarks by $\mathcal{Z} = \{\mathbf{z}_k\}_{k=1}^P$ and $\mathcal{X} = \{\mathbf{x}_k\}_{k=1}^S$ where $\mathbf{z},\mathbf{x}\in \mathds{R}^2$ are the 2D locations of the primary and secondary landmarks in an image, respectively, and $P$ and $S$ are the number of primary and secondary landmarks, respectively. $\mathcal{D}_{\mathcal{Z}}=\{\mathbf{I}_i,\mathcal{Z}_i\}_i$ and $\mathcal{D}_{\mathcal{X}}=\{\mathbf{I}_i,\mathcal{X}_i\}_i$ are the labeled primary and secondary landmark datasets where $\mathbf{I}_i$ is the $i^{\rm th}$ image. Since the secondary landmarks are difficult to annotate, the size of the secondary landmark dataset is substantially (at least an order of magnitude) smaller than that of the primary landmark dataset, i.e., $|\mathcal{D}_{\mathcal{Z}}|\gg|\mathcal{D}_{\mathcal{X}}|$. We denote the total dataset by $\mathcal{D} = \mathcal{D}_\mathcal{Z} \cup \mathcal{D}_\mathcal{X} \cup \mathcal{D}^U_\mathcal{X} = \mathcal{D}^{\rm L} \cup \mathcal{D}^U_\mathcal{X}$ where $\mathcal{D}^U_\mathcal{X}$ is the dataset of the unlabeled multiview images of the secondary landmarks, and $\mathcal{D}^{L}=\mathcal{D}_\mathcal{Z} \cup \mathcal{D}_\mathcal{X}$ is the labeled dataset of the primary and secondary landmarks. We assume that the multiview cameras are static, and their intrinsic and extrinsic parameters are pre-calibrated.

\subsection*{3D Secondary Landmark Prediction}
We learn a visual representation by predicting the 3D locations of the secondary landmarks. Specifically, we learn a new function that encodes the spatial relationship between the primary and secondary landmarks, i.e., 
\begin{align}
    \mathcal{X}_{\rm 3D} = f(\mathcal{Z}_{\rm 3D}), \label{Eq:predictor}
\end{align}
where $\mathcal{Z}_{\rm 3D} = \{\mathbf{Z}_k\}_{k=1}^P$ and $\mathcal{X}_{\rm 3D}  = \{\mathbf{X}_k\}_{k=1}^S$ are the sets of the 3D primary and secondary landmarks, respectively, and $\mathbf{Z},\mathbf{X}\in\mathds{R}^3$ are the 3D locations of the primary and secondary landmarks.

As an input to Equation~(\ref{Eq:predictor}), we reconstruct the 3D primary landmarks from two views:
\begin{align}
    \mathbf{Z}_k = \phi(\mathbf{z}_k^i, \mathbf{z}_k^j, \Pi_i, \Pi_j),
\end{align}
where $\mathbf{z}_k^i$ is the $k^{\rm th}$ 2D primary landmark in the $i^{\rm th}$ view image, and $\phi$ reconstructs the 3D primary landmarks from two views. $\Pi_i:\mathds{R}^3\rightarrow \mathds{R}^2$ is the projection function onto the $i^{\rm th}$ image that encodes its camera intrinsic and extrinsic parameters. In practice, we leverage a direct linear transform~\cite{hartley:2004} to linearly triangulate the 3D landmarks, which is differentiable.

Instead of using the triangulated coordinates directly, we use a coordinate normalization to learn a geometrically coherent representation for the secondary landmark predictor $f$. Given a triangulated primary landmark set $\mathcal{Z}_{\rm 3D}$, we perform Procrustes analysis~\cite{sorkine2017least} to align the primary landmarks:
\begin{align}
    \widehat{\mathbf{Z}} = s\mathbf{R}\mathbf{Z} + \mathbf{t},
\end{align}
where $s\in \mathds{R}$, $\mathbf{R}\in SO(3)$, and $\mathbf{t}\in \mathds{R}^3$ are the scale, rotation, and translation, estimated by a Procrustes analysis. In practice, we use the spine and shoulder limbs to define the coordinate system of a pose, i.e., the spine limb as the $x$-axis, and the right shoulder limb to the $y$-axis where the coordinate is scaled such that the spline limb has a unit length.

\subsection*{Semi-supervised Multiview Loss}
We denote the landmark detector for the $k^{\rm th}$ landmark by $\Psi_k:\mathds{I}\rightarrow \mathds{R}^2$ that takes an image $\mathbf{I}$ and outputs the 2D location of the landmark, i.e., $\mathbf{x}_k = \Psi_k(\mathbf{I})$. We model the landmark detector by decomposing it into the feature extractor that learns a common visual representation across landmarks, and the landmark localizer the finds the 2D location given the visual representation. 

The feature extractor $\Phi:\mathds{R}^2\times \mathds{I} \rightarrow \mathds{R}^n$ is a function that extracts a visual feature from an image where $\mathds{I}=\mathds{R}^{3\times H \times W}$ is the image range ($H$ and $W$ are its height and width, respectively), and $n$ is the dimension of the visual feature, i.e., $\Phi(\mathbf{x}, \mathbf{I})$ is the visual feature (e.g., the penultimate layer of a 2D landmark detector) at $\mathbf{x}\in \mathds{R}^2$ on the image $\mathbf{I} \in \mathds{I}$. On the other hand, the landmark localizer 
$\psi_k:\bunderbrace{\mathds{R}^n\times \cdots \times \mathds{R}^n}{WH}\rightarrow \mathds{R}^2$ estimates the 2D location of the $k^{\rm th}$ landmark from the visual representations, i.e., $\Psi_k = \psi_k\circ \{\Phi(\mathbf{x}, \mathbf{I})\}_{\mathbf{x}}, \forall \mathbf{x} \in [0,\cdots,H-1]\times [0,\cdots,W-1]$. 

In practice, we design the feature extractor with deep convolutional layers to learn a high level and generic representation for all landmarks while the landmark localizer is modeled by shallow layers that are responsible for the landmark classification. 

We minimize the following objective to jointly learn the landmark detector $\Psi$ and predictor $f$ using both labeled and unlabeled multiview images:
\begin{align}
    \mathcal{L}(\boldsymbol{\theta}_\Phi, \boldsymbol{\theta}_f, \boldsymbol{\theta}_\psi) = \sum_{(\mathbf{I}_i,\mathbf{I}_j)\in\mathcal{D}} \mathcal{L}_{\rm U}(\mathbf{I}_i,\mathbf{I}_j) + \lambda_{\rm L} \sum_{\mathbf{I}\in\mathcal{D}^{L}} \mathcal{L}_{\rm L}(\mathbf{I},\overline{\mathcal{X}}, \overline{\mathcal{Z}}),\nonumber
\end{align}
where $\mathcal{L}_{\rm U}$ and $\mathcal{L}_{\rm L}$ are the losses for the unlabeled and labeled data, and $\lambda_{\rm L}$ controls the balance between two losses. $\boldsymbol{\theta}_\Phi, \boldsymbol{\theta}_f, \boldsymbol{\theta}_\psi$ are the weights parametrizing the functions $\Phi, f, \psi$. $\overline{\mathcal{X}}$ and $\overline{\mathcal{Z}}$ are the ground truth primary and secondary landmarks, respectively. $(\mathbf{I}_i,\mathbf{I}_j)$ is a pair of synchronized multiview images.

\begin{figure*}[t]
\begin{center}
\includegraphics[width=1\textwidth]{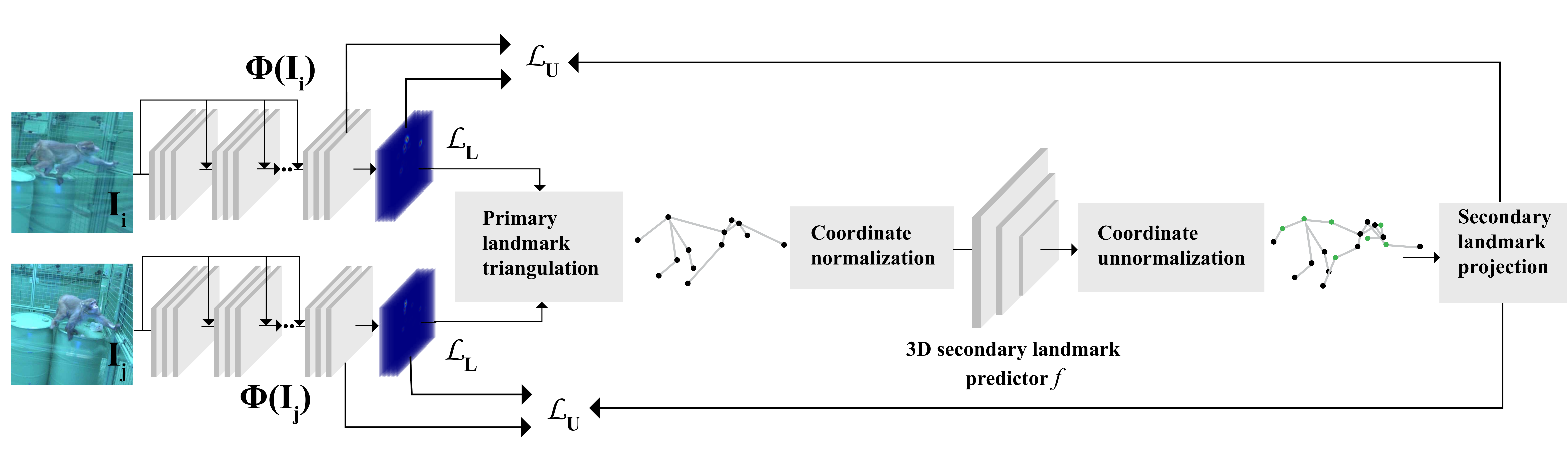} 
\end{center}
\vspace{-5mm}
\caption{Overall architecture of the proposed framework. We design twin networks to predict the 2D primary and secondary landmarks from two views. The predicted primary landmarks are triangulated to form the 3D primary landmarks. With normalization through Procrustes analysis, the 3D secondary landmarks can be predicted by 3D landmark predictor. We use the predicted secondary landmarks to evaluate geometric error (reprojection error) and equivariance measures (feature correlation). We minimize the label and unlabel losses ($\mathcal{L}_{\rm L}$ and $\mathcal{L}_{\rm U}$).}
\label{fig:short}
\end{figure*}

$\mathcal{L}_{\rm U}$ measures the geometric consistency of the secondary landmarks between a pair of views and uses contrastive learning:
\begin{align}
    \mathcal{L}_{\rm U}(\mathbf{I}_i,\mathbf{I}_j) &= \sum_{k}\left\|\Psi_k(\mathbf{I}_i)-\Pi_i(\mathbf{X}_k)\right\|^2  + \left\|\Psi_k(\mathbf{I}_j)-\Pi_j(\mathbf{X}_k)\right\|^2\nonumber\\
    &-\sum_{k} \langle\Phi_k(\Pi_i(\mathbf{X}_k), \mathbf{I}_i),\Phi_k(\Pi_j(\mathbf{X}_k);\mathbf{I}_j)\rangle \nonumber\\
    &+\sum_{k\neq l} \langle\Phi_k(\Pi_i(\mathbf{X}_k), \mathbf{I}_i),\Phi_l(\Pi_i(\mathbf{X}_l);\mathbf{I}_i)\rangle,
\end{align}
where $\mathbf{X}_k$ is predicted by $f$ from the triangulation of the primary landmarks via Equation~(\ref{Eq:predictor}). $\langle \cdot, \cdot \rangle$ measures the normalized cross-correlation between two vectors. Note that the loss is agnostic to the labels of the secondary landmarks where the total data $\mathcal{D}$ including the unlabeled multiview image pairs can be used to learn the detector and predictor jointly. 

The loss for the unlabeled data encodes the complementary relationship of geometry and visual semantics. The first two terms ensure the geometric consistency by minimizing the reprojection error, i.e., the detected secondary landmarks from $\Psi(\mathbf{I})$ must align with the projection of the predicted 3D landmarks $\Pi(\mathbf{X})$. The third term enforces contrastive learning across views, i.e., the visual representation of the corresponding landmarks must be view invariant. In addition, the last term enforces the uniqueness of the visual features across landmarks, e.g., the visual feature of elbow must be sufficiently different from that of wrist. These consistency and uniqueness measures facilitate self-supervised learning, i.e., the geometric consistency allows us to precisely localize the landmarks via a consensus of predictions, and the equivariance enforces view-invariance in learning the visual representation for the landmark detector. As a result, the feature correlation of the same landmark across views is high while that of different landmarks within the same view is low as shown in Figure~\ref{Fig:supp1b}.

Jointly learning visual representation and 3D landmark prediction allows us to apply multiview supervision where visual information from one view can be transferred to the other views through multiview geometry. This results in utilizing a large amount of unlabeled data of the secondary landmarks in conjunction with a small set of labeled data. 

$\mathcal{L}_{\rm L}$ is the loss for the labeled data, which can be defined as: 
\begin{align}
    \mathcal{L}_{\rm L}(\mathbf{I}, \overline{\mathcal{X}}, \overline{\mathcal{Z}}) = \sum_{k=1}^S \|\overline{\mathbf{x}}_k - \Psi_k(\mathbf{I})\|^2 + \sum_{k=S+1}^{S+P} \|\overline{\mathbf{z}}_k - \Psi_k(\mathbf{I})\|^2, \nonumber
\end{align}
where $\overline{\mathbf{z}}$ and $\overline{\mathbf{x}}$ are the ground truth primary and secondary landmarks, respectively. The first $S$ outputs from the landmark detector $\Psi$ are the secondary landmarks and the next $P$ outputs are the primary landmarks.

\subsection*{Network Design and Implementation}
We design a neural network that facilitates jointly learning the landmark detection $\Psi$ and 3D landmark prediction $f$, by leveraging multiview supervision as shown in Figure~\ref{fig:short}. Given a pair of images from different views at the same time instant, the twin landmark detectors that share the weights produce the predictions of the primary and secondary landmarks in the form of heatmaps (i.e., probability of the landmarks). We use the 6 stage convolutional pose machine~\cite{wei2016cpm} as a pose estimator that takes as input an image with the size of 368$\times$368 and outputs a set of heatmaps with the size of 46$\times$46$\times(P+S+1)$ including the background. The soft-argmax on the heatmaps is used to estimate the coordinates of the landmarks. This landmark detector is complementary to other pose detectors~\cite{newell:2016,toshev:2014,cao2017realtime}, and our method is agnostic to the choice of networks. The primary landmarks from these two views are triangulated to form the 3D primary landmarks, and transformed to the normalized coordinate. The visual features are extracted from the penultimate layer of the pose estimator at the 2D location of the predicted primary landmarks. We implement the 3D landmark predictor $f$ using a multi-layer perceptron with three hidden layers that predicts the 3D secondary landmarks. These predicted secondary landmarks are projected onto each view for geometric and semantic consistency (contrastive learning). In practice, we pretrain the landmark detector and 3D predictor using the labeled data $\mathcal{D}^{L}$, and then refine them by minimizing the overall loss $\mathcal{L}$ with the total data $\mathcal{D}$ that includes labeled and unlabeled multiview data in an end-to-end fashion. In training, we use batch size 10, learning rate $10^{-4}$, and learning decay rate 0.8 with 2000 steps. We use the ADAM optimizer~\cite{Kingma2014} of TensorFlow with a single NVIDIA GTX 2080Ti. The value of $\lambda_L$ has been set to 10.

\subsection*{Datasets}
We evaluate our method on the following datasets.

\noindent\textbf{OpenMonkeyPose} is a large landmark dataset of rhesus macaques captured by 62 synchronized multiview cameras. It consists of nearly 200k labeled images with four macaque subjects that freely move in a large cage while performing foraging tasks. 13 primary landmarks are annotated by the crowd-workers, including nose, head, neck, shoulders, hands, hip, knees, feet, and tail. In addition to the primary landmarks, we manually annotate 14k images of the secondary landmarks (elbows, ears, spine and a mid-point of the tail) by incorporating 3D reconstruction from synchronized images. We train our secondary landmark detector using 14k labeled images and 160k unlabeled multiview images.

\noindent\textbf{Human3.6M} is a human pose dataset captured by 4 high definition cameras that includes 7 subjects performing a variety of activities such as eating, greeting, sitting, and walking. The data consists of 32 annotated joints per image which include nose, head, neck, ears, shoulders, elbows, wrists, hip, thorax, spine, knees, and feet. We make use of 14 joints as primary landmarks, including nose, head, neck, shoulders, hands, hip, right-hip, left-hip, knees, and feet, and three joints as secondary landmarks, including elbows and spine. We train our model using 30k labeled images from the subjects 1 and 5, and 180k unlabeled images from the subjects 6, 7 and 8.

\noindent\textbf{DeepFly3D} contains a large number of landmark annotations of Drosophila: adult flies captured by seven synchronized multiview cameras. The dataset comprises nearly 1M images with 10 different subjects captured in the course of four varying experiments. The data consists of 38 landmark locations, which include, five on each limb\textemdash the thorax-coxa, coxa-femur, femur-tibia, and tibia-tarsus joints as well as the pretarsus, six on the abdomen\textemdash three on each side and one on each antenna. We make use of 12 joints as primary landmarks, including the thorax-coxa, coxa-femur, femur-tibia, and pretarsus for the left-hand side limbs of the fly. The three tibia-tarsus joints on the left-hand side limbs are used as secondary landmarks. We train our model on 3k labeled images and 35k unlabeled images.